# A Robust System for Facial Emotions Recognition Using Convolutional Neural Network

Faisal Ghaffar[1], Imad Ali[2], Sarwar Khan[3], Wasim Ahmad[3], Khursheed Ali[4],

[1]System Design Engineering Department, University of Waterloo, Canada

[2]Department of Computer Science, University of Swat, KP, Pakistan

[3]Institute of Information Sciences, Academia Sinica, Taiwan &, Department of Computer Science, National ChengChi University, Taiwan

[4]Department of Computer Science, Insititute of Bussiness Administration, Sukker, Sindh, Pakistan

## ABSTRACT

Facial expressions vary from person to person, and the brightness, contrast, and resolution of every random image are different. This is why recognizing facial expressions is very difficult. This article proposes an efficient system for facial emotion recognition for the seven basic human emotions (angry, disgust, fear, happy, sad, surprise, and neutral), using a convolution neural network (CNN), which predicts and assigns probabilities to each emotion. Since deep learning models learn from data, thus, our proposed system processes each image with various pre-processing steps for better prediction. Every image was first passed through the face detection algorithm to include in the training dataset. As CNN requires a large amount of data, we duplicated our data using various filters on each image. Pre-processed images of size 80*100 are passed as input to the first layer of CNN. Three convolutional layers were used, followed by a pooling layer and three dense layers. The dropout rate for the dense layer was 20%. The model was trained by combining two publicly available datasets, JAFFE and KDEF. 90% of the data was used for training, while 10% was used for testing. We achieved maximum accuracy of 78.1 % using the combined dataset. Moreover, we designed an application of the proposed system with a graphical user interface that classifies emotions in real-time.

## I. INTRODUCTION

HUMAN facial expressions have a significant role in involvement, engaging, or communicating with each other. A machine with high accuracy and powerful expression recognition intelligence will better interpret human emotions and interact more naturally [1, 2]. A face detection system is used to detect faces in an image/video and then processes those images to obtain visible facial features, while the emotion recognition system detects faces and recognizes the respective emotion [3]. Many researchers have studied and conducted experiments to produce better results and, ultimately, a system that recognizes emotions. In real-world applications such as commercial call centers, the system of emotions screening during the interview, screening behavior of a student in a class or MOOCs, and affect-aware game development will significantly benefit from such a system. Thus, developing an efficient, low complexity, and robust system that can provide a probabilistic score for each emotion class is crucial. Also, developing an application for facial emotion detection from live video and still images is essential for many real-world scenarios.

Many face detections have been studied in computer vision with massive datasets and complex models. For example, Vaillant et al. [4] used neural networks for face detection. They trained a convolutional neural network and scanned the image with a window to locate the face. Rowley et al. [5] developed a connected neural network to detect a face in the image. Publicly available benchmarks such as Face Detection Database and Benchmark [6], wilder Face-Face Detection Benchmark [7], and PASCAL FACE [8] also have contributed to the advancement of face detection problems. Generally, the latest face detection algorithms can be classified into these categories: cascade-based algorithms [9, 10], part-based algorithms [8, 11], channel feature-based algorithms [12, 13], and neural network-based algorithms [14, 15]. Extensive work has also been done on facial expression recognition. Many researchers have worked on the systems and used Facial action coding [16-18]. Hidden Markov model neural network-based models have also been proposed in several studies for facial emotion detection [19]. However, these approaches primarily focus on facial investigation keeping background intact, and hence built up a lot of unnecessary and misleading features that confuse the CNN training process. Moreover, these approaches use complex models that cause huge delays when turning into real-world facial recognition applications.

This article proposed an efficient and low-complexity system to predict face emotion. The proposed system assigns a probabilistic score to each of the seven essential facial expression classes: angry, disgust, fear, happy, sad, surprise, and neutral, using a convolutional neural network (CNN) that does not build up unnecessary and misleading features in the training process. First, the proposed system detects the face in an image and then estimates the face area for extraction. Since deep learning models learn from data, thus, the proposed system processes each face with diverse preprocessing steps so that our neural network may recognize those features with high accuracy. The image having the highest probability is assigned to the respective emotion class. The model is trained on the two publicly available limited and challenging datasets, where the proposed model achieved an accuracy of 78.1%, which is higher than the baseline models. Finally, we developed our application for facial emotion detection from live video and still images using that model. The contributions of this article are as follows:

* Corresponding authors:
**Faisal Ghaffar**
System Design Engineering Department, University of Waterloo, Canada, Email: faisal.ghaffar@uwaterloo.ca

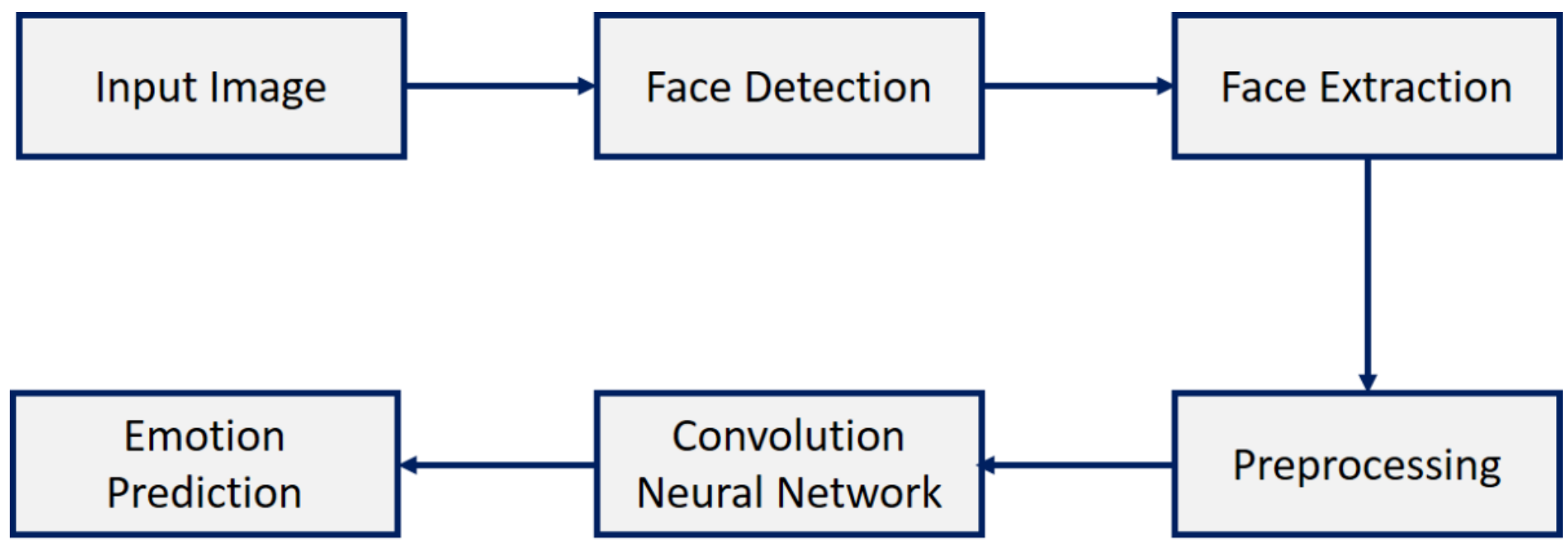

**Figure 1:** The block diagram of the proposed system.

1. We proposed an efficient system to predict face emotion using a convolution neural network, which assigns probabilities to each emotion.

2. Since deep learning models learn from data, thus, our proposed system processes each image with diverse preprocessing steps for better prediction.

3. We designed an application of the proposed system with a graphical user interface that classifies emotions in real-time.

The rest of the article is organized as follows: Section 2 presents the proposed system for emotion recognition. Section 3 evaluates the proposed system and presents the datasets and results. Section 4 presents the graphical user interface of the proposed system. Section 5 concludes this article.

## II. THE PROPOSED SYSTEM

Figure 1 shows the block diagram of the proposed systems. First, in a given image, the face is detected, and then it is extracted. The extracted face goes under extensive preprocessing and is then fed to the convolution neural network (CNN), assigning emotion probability to each emotion. The image with the highest probability is given to the respective emotion class. These are discussed in detail in the following.

### *A. Face Detection*

Many algorithms are proposed for face detection, like, Haar cascade, HOG +, Support Vector Machine, or deep learning models. Our system utilized the algorithm proposed in [20], which is faster and can work in real-time. This algorithm finds the locations of 68 (x, y) coordinates. Each of these represents a location of the region of interest on the face. Those 68 points are shown in Figure 2. This algorithm uses a training set of labeled facial points on an image, which is manually labeled. Those labels are (x, y) coordinates that specify regions of the face. Points 1-17 define the jawline, 18-22 specify the left eyebrow, 23-27 specify the right eyebrow, 37-42 encircle the left eye region, 43-48 encircle the right eyes region, 28-36 specify the region of the nose, 49-60 specify the outer lip area, while 60-68 specifies inner lip structure. This data is then used for the training of a model of regression trees to find these facial landmark points directly from the pixel intensities. There is no feature extraction in this method; hence the detection is very fast, i.e., in real-time and with high accuracy and quality. We do not need ear regions and regions above mid of the forehead because emotion mainly deals with the eyes, nose, eyebrows regions, and some facial regions. So, this face detection algorithm gives us the exact facial areas we are interested in. Figure 3 shows some results of our detected faces and areas according to the 68 points model.

### *B. Face Extraction*

After detecting the faces, the next step is to extract that face region from the original image and drop the rest. We first extracted 1-27 points locations from 68 points facial landmarks. These 1-27 points cover the overall face, which deals with facial expressions. To find the minimum enclosed structure of a set, we use a convex hull. The convex hull finds the minimum possible enclosed structure of the set of points. Thus, we applied a convex hull to those 27 points to create an enclosed structure boundary of the face. Figure 4 (Left) shows a convex hull of 0-27 points. To make a mask to extract the face, we fill the convex using a convex polygram. The mask size is the same as that of the original image. Figure 4 (Right) shows the mask obtained after convex filling.

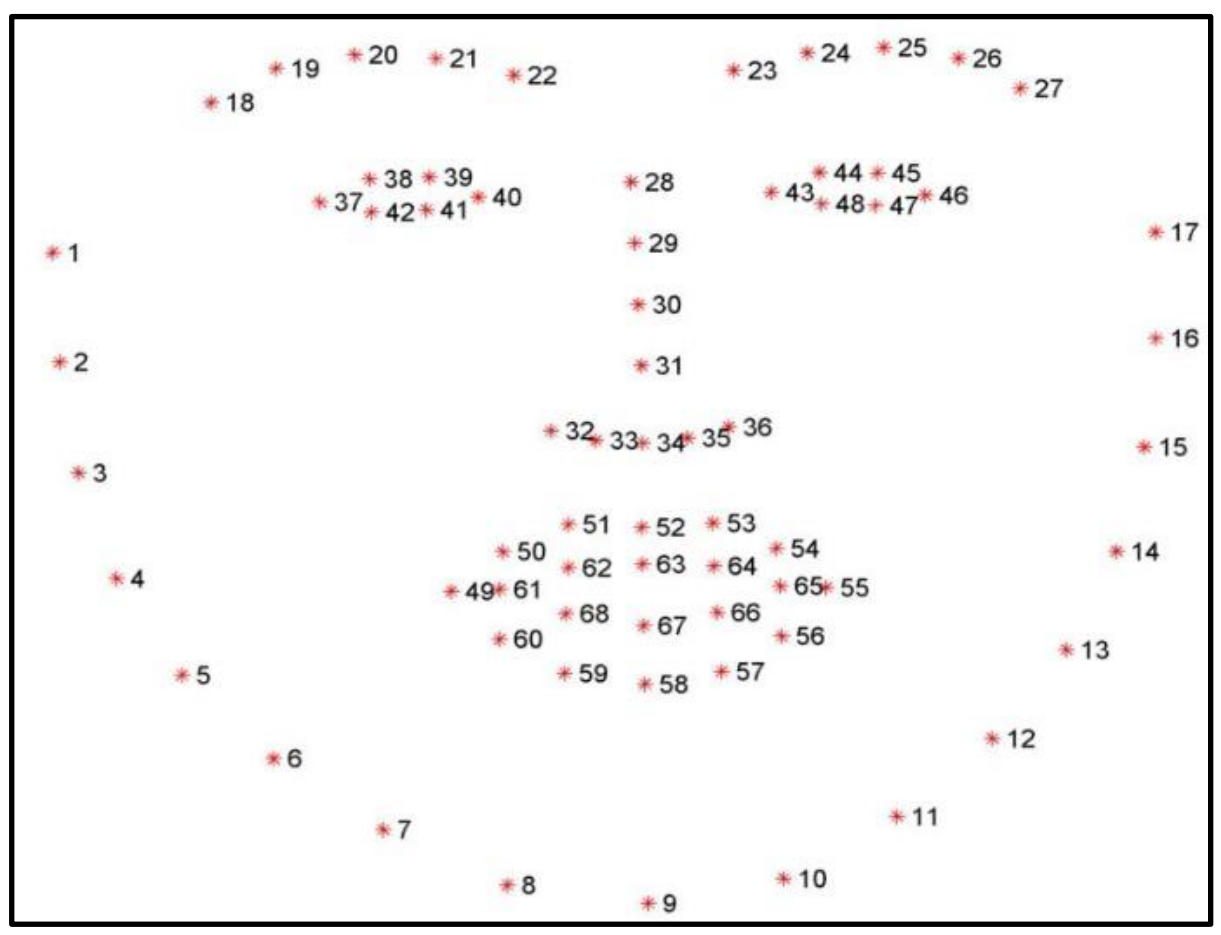

**Figure 2**: Visualizing the 68 facial landmark coordinates [20].

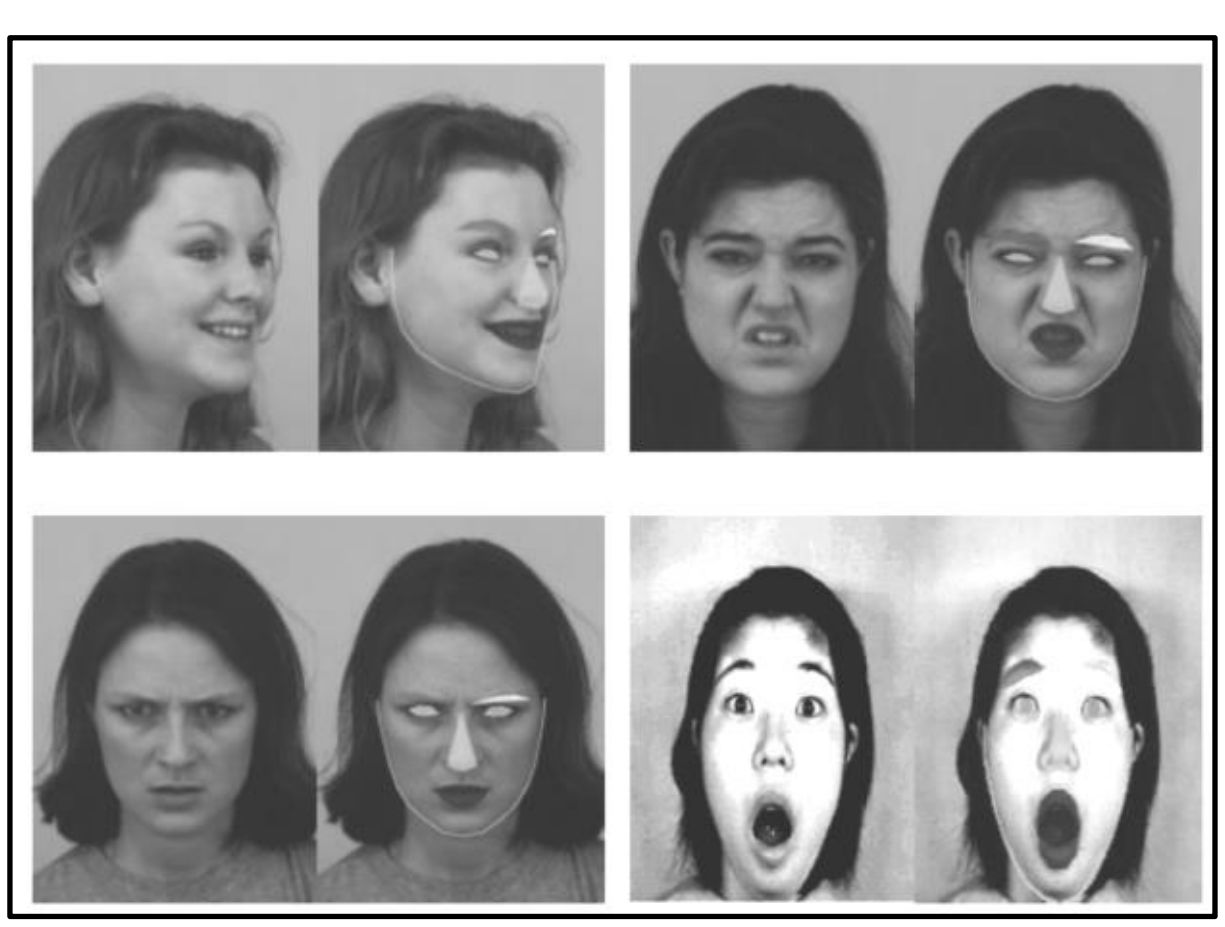

**Figure 3**: Face detection and face area estimation

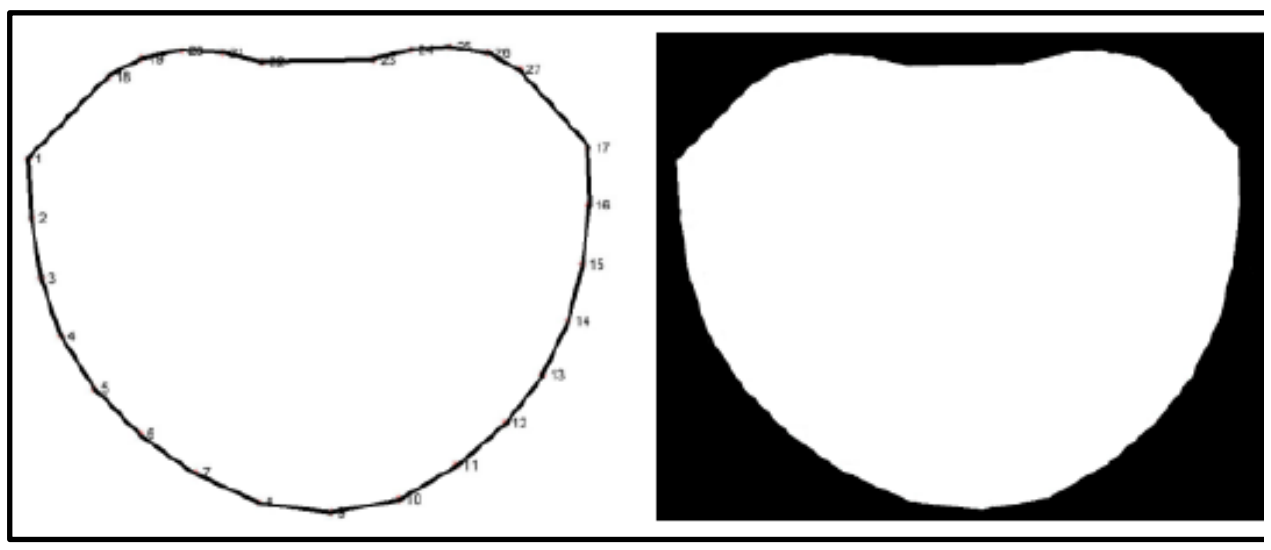

**Figure 4: (Left)** Convex hull of 0-27 facial landmarks, **(Right)** Mask obtained after filling the convex hull

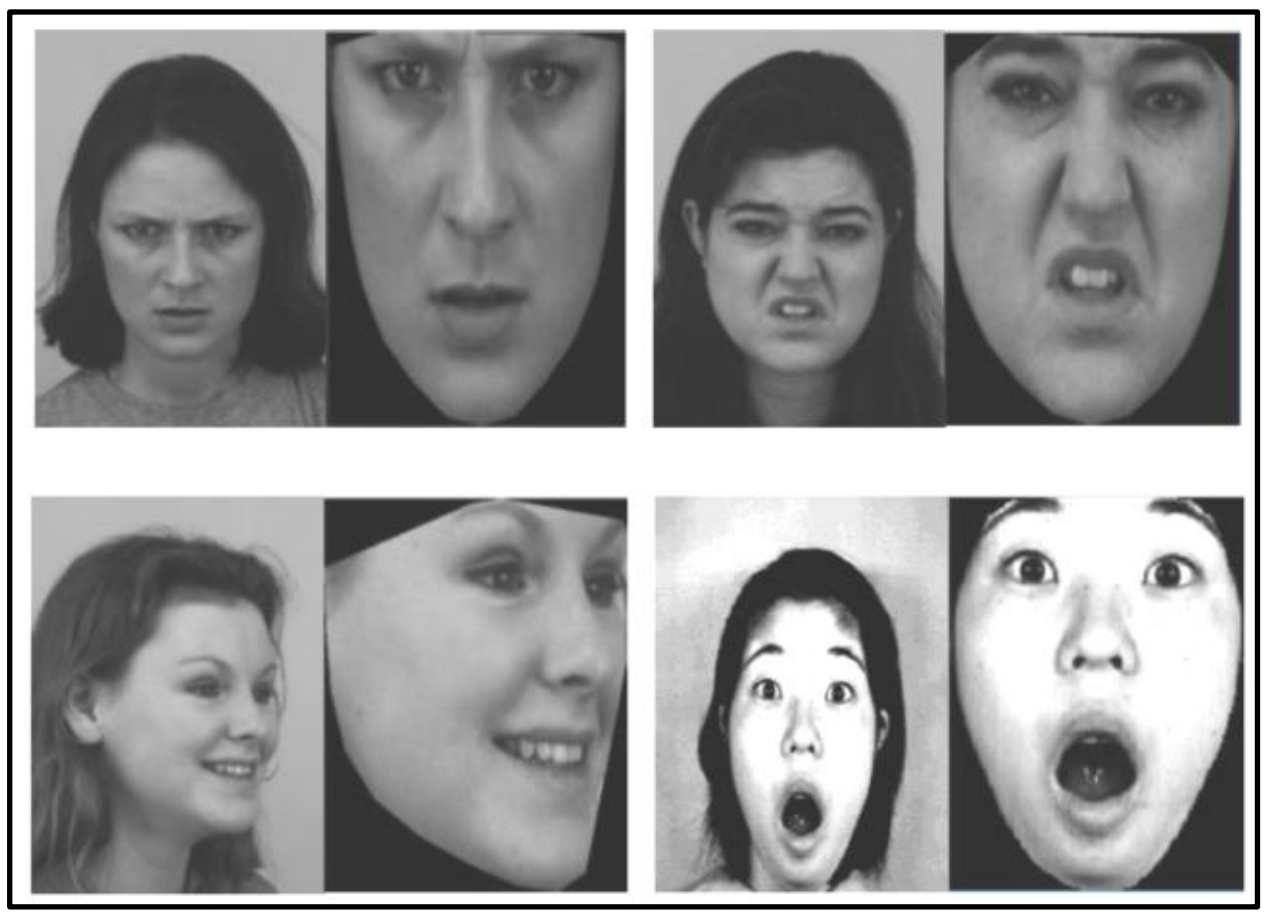

**Figure 5:** Face extraction process

### *C. Preprocessing*

The mask was then applied to images to extract the face area. Figure 5 shows the extracted face images. Since deep learning learns from data, thus, it is crucial to feed proper data to the model. Thus, we apply a series of preprocessing before feeding the data into CNN. First, we used histogram equalization on the cropped face images for intensity normalization and contrast enhancement. We then used the bilateral filter to remove noise while preserving the edges. A bilateral filter combines both domain and range filtering. Figure 6 shows how the bilateral filter removes noise and preserves high-frequency edges. Then we apply convolutional 2D filter with a kernel of ([-1, -1, -1], [-1, 9, -1], [-1, -1, -1]). We then resized all images to 80*100. An array of training and test images were made to pass it into the CNN model.

### *D. CNN Architecture*

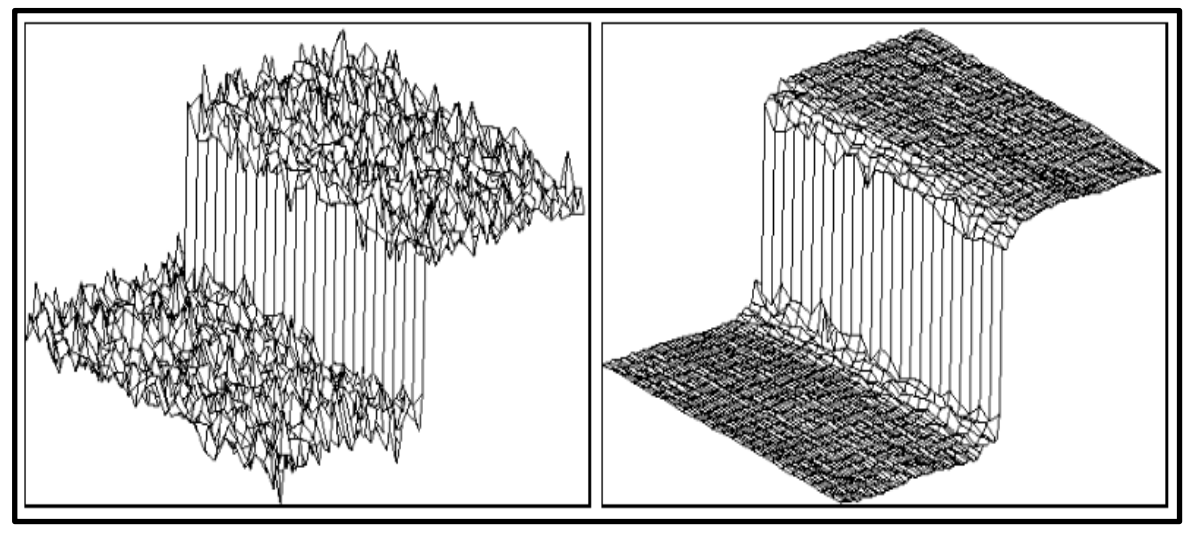

**Figure 6**. Bilateral filter noise removal and edge-preserving

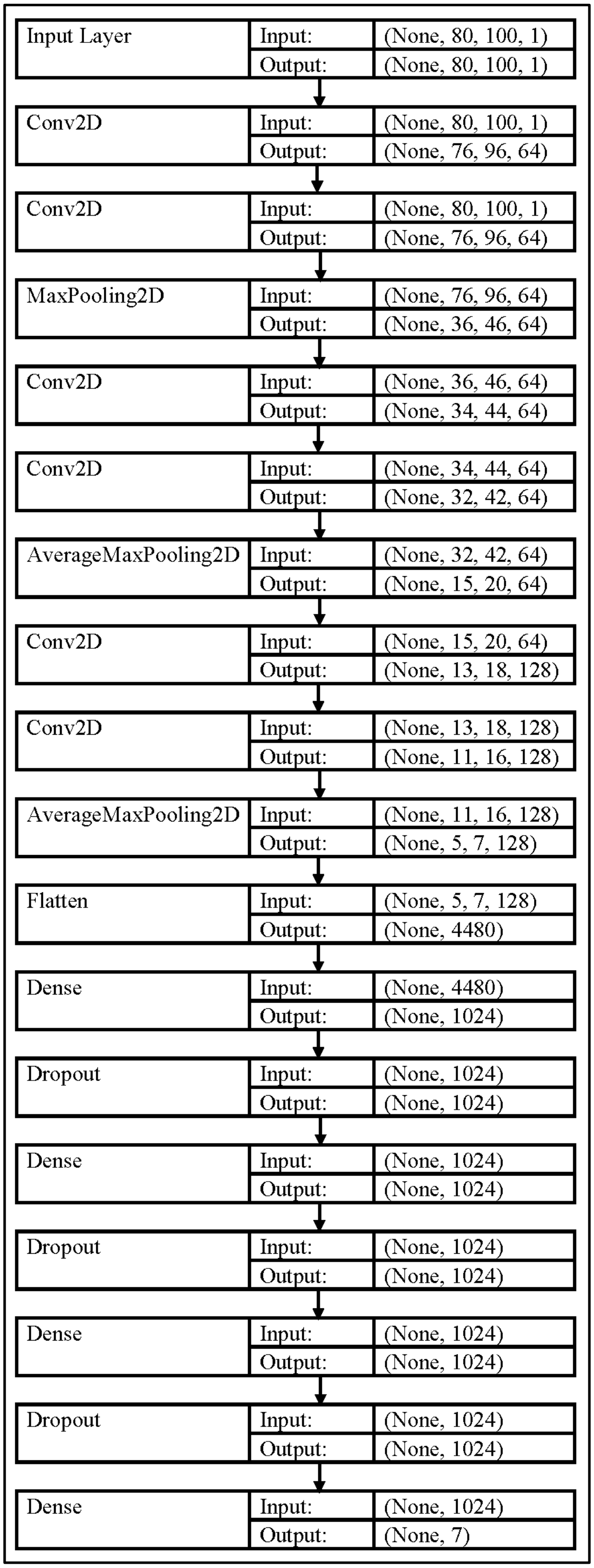

**Figure 7**: The proposed CNN architecture.

The proposed CCN model has five convolutional layers, one max pooling layer, two average pooling layers, and three dense layers. Max pooling chooses the maximum response, while the average filter calculates the average of the responses. The dense layers contain a dropout of 20%. This dropout in connected layers prevents the model from overfitting. We have train data of shape (14210,100,80) and test data of shape (1580,100,80). The shape tells us that there are 14210 training instances and 1580 test instances, each of which has a dimension of 80*100, the size of our image. The input layer accepts 80*100-dimension images and passes the data of the same dimension to the first convolutional layer. The convolutional layer has 64 filters and a kernel of size (5,5) and has the activation function ReLu. The layer gives an output (76,96,64), which is passed to the max pooling layer with a pool size of (5,5) and strides of (2,2). A result of size (36,46,64) is obtained, then passed to two convolutional layers in series. Both have 64 filters, kernel size (3,3), and a relu activation function. The output of the second convolutional layer in the series has a size of 32,42,64. An average pooling of pool size (3,3) and strides (2,2) is connected next in the series. The output size of this average pooling layer is (15,20,64). It is further associated with two consecutive convolutional layers with 128 filters and kernel size (3,3) and has the activation function relu. The output of these last convolutional layers has the size of (11,16,128). To this output, average pooling with the same pool size and stride as the earlier one is applied. The output of this layer has a size (5,7,128), which is then flattened 5*7*128 = 4480, passed to three fully connected layers with 1024 filters each.

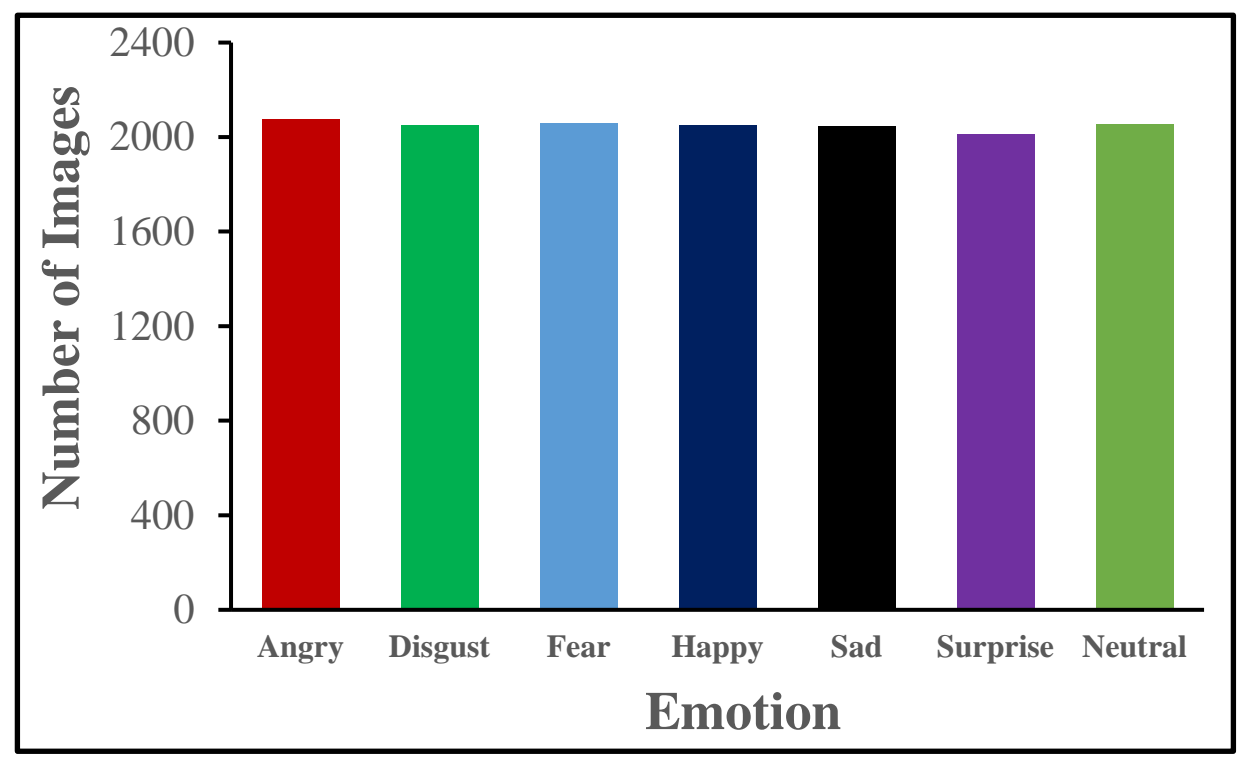

(a) Training data

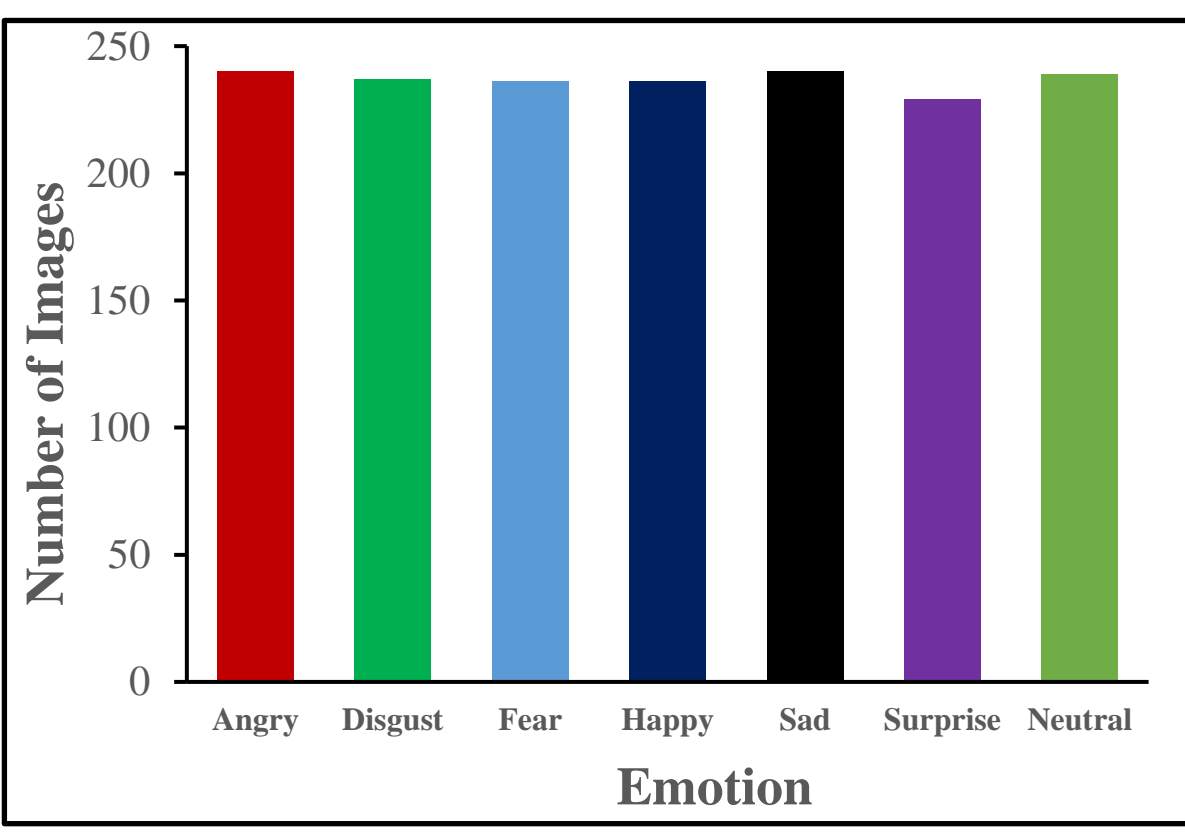

(b) Testing data

**Figure 8**: The emotion dataset with (a) training and (d) testing data

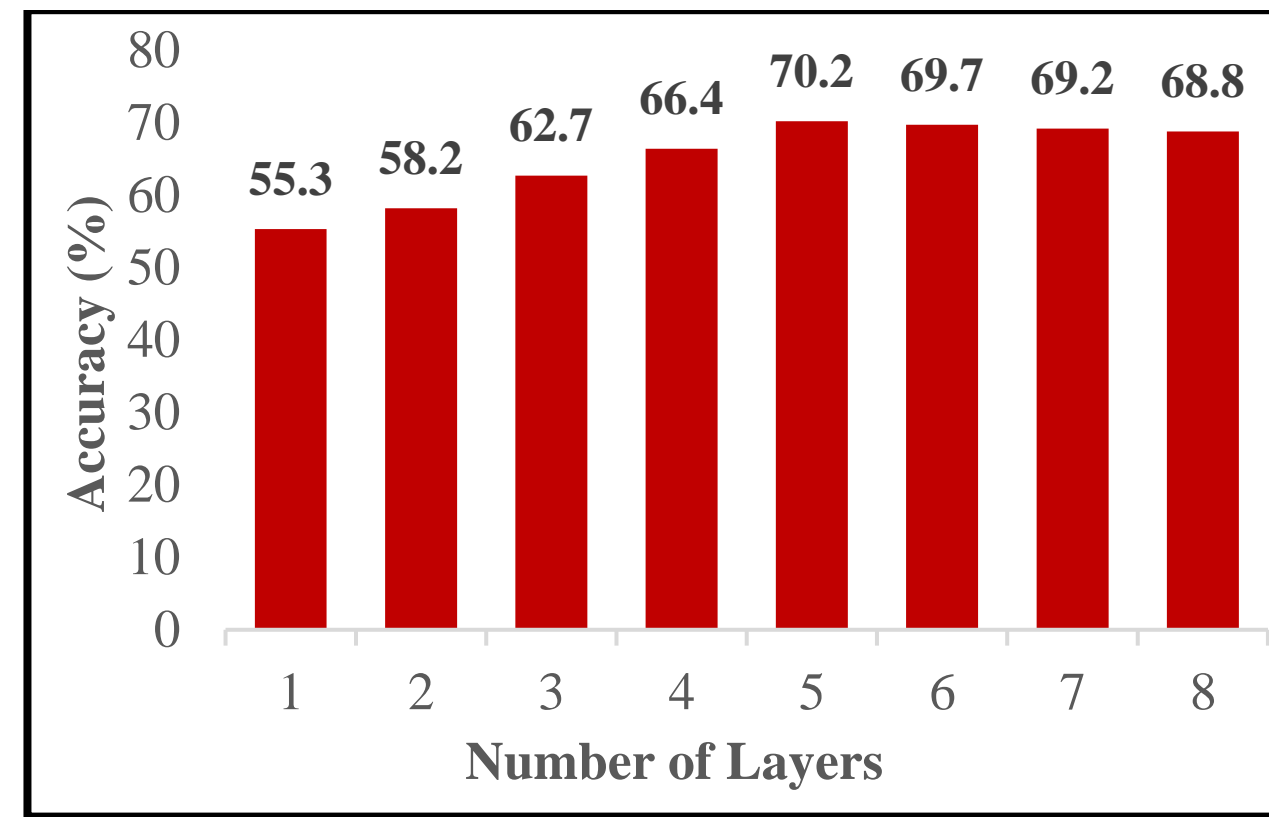

**Figure 9:** Accuracy vs. different number of layers.

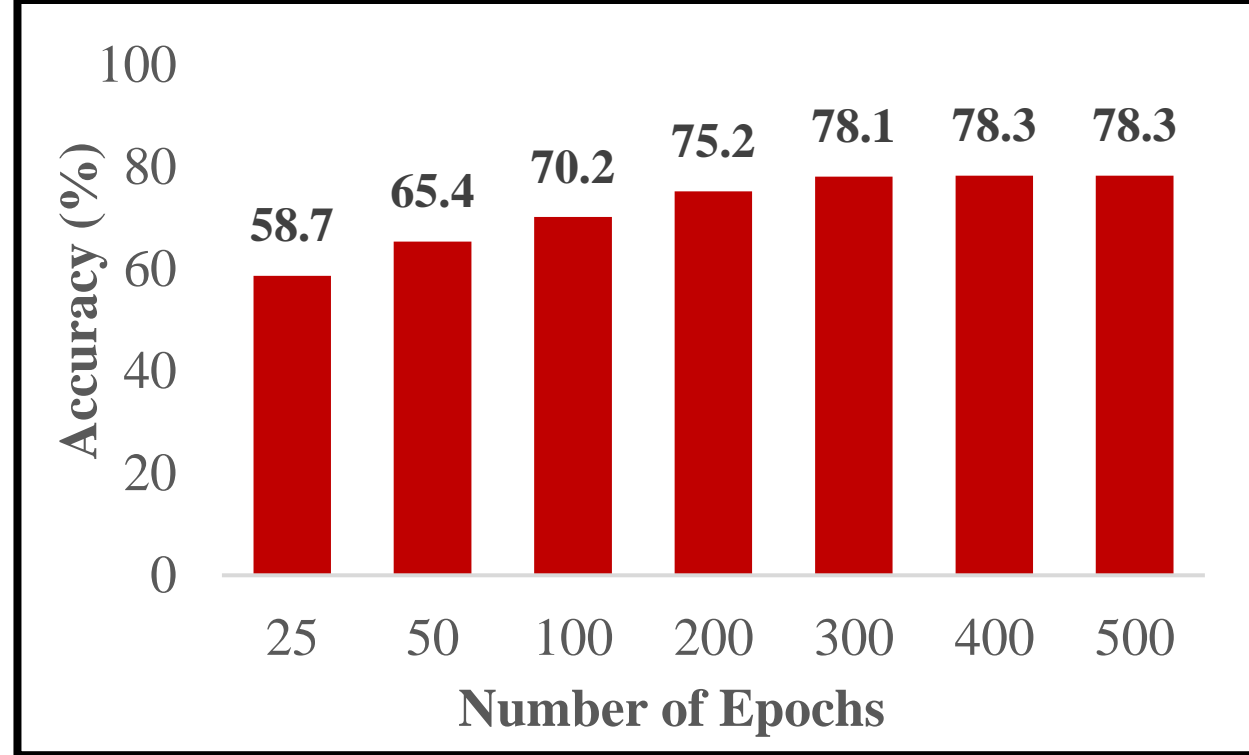

**Figure 10**: Accuracy vs. diverse number of epochs

The dropout rate of each layer is 0.2. Finally, the layer is dense with the activation function SoftMax, which gives the output of size 7. These seven instances represent the probability of each class. In the convolutional layer activation function, relu is used instead of sigmoid because it reduces the likelihood of gradient becoming too small and almost disappearing, and also relu results in sparse representation while sigmoid results in dense representation. The earlier is much better than the latter. The proposed CNN architecture is shown in Figure 7.

## III. EVALUATION

This section presents the dataset details and evaluates our proposed systems.

### *A. Dataset*

We have combined two datasets Japanese Female Facial Expression dataset, and Karolinska Directed Emotional Faces Dataset [21, 22]. The Japanese Female Expression dataset contains 213 static images of 10 models. The images are grayscale, and the resolution is 256*256. All the images are posted. The dataset includes 4900 pictures from different models. All images are grayscale with resolution 572*762. The dataset contains images of 70 models. There is an equal distribution of males and females. The images are taken in the same lighting conditions with no makeup, glass, or earrings.

Each model is photographed from 5 different views (frontal, entire left, half left, full right, and half right). The mouth and eyes are positioned on a grid at a specific place, and then images are cropped at a specified resolution. The age of the model is in between 20 to 30 years. Both datasets contain seven facial expressions (angry, disgust, fear, happy, sad, surprise, neutral), which we labeled as (0, 1, 2, 3, 4, 5, 6), respectively. We have combined both datasets according to each emotion label. After preprocessing and duplicating data in these datasets data were classified into training and testing sets. The testing data is almost 11% of that of the training data. Training data contain 14200 images, and test data contain 1580 images. Images whose faces were not detected in the preprocessing stage were removed from the dataset as our primary focus is on expression recognition rather than face detection problems. The dataset is almost balanced. Statistics of each emotion set of final datasets are shown in Figure 8.

**Table 1:** Precision, Recall, F1-Score, and Support

| Emotion | Precision | Recall | F1-Score | Support |
|---|---|---|---|---|
| 0 | 0.84 | 0.56 | 0.67 | 230 |
| 1 | 0.82 | 0.86 | 0.84 | 225 |
| 2 | 0.60 | 0.60 | 0.60 | 225 |
| 3 | 0.90 | 0.88 | 0.89 | 225 |
| 4 | 0.79 | 0.83 | 0.81 | 230 |
| 5 | 0.87 | 0.84 | 0.85 | 220 |
| 6 | 0.67 | 0.85 | 0.75 | 225 |
| Avg. | 0.78 | 0.77 | 0.77 | 1580 |

### B. Results

We trained the network with a batch size of 100 and 100 epochs for layers in {1, 2, 3, 4, 5, 6, 7, 8} Epochs are one complete pass-over training set instance. We varied the number of CNN layers from 1 to 8 to find the best accuracy. We found that the best accuracy in our setting is obtained on five layers. The execution time increased with the number of layers without improving the accuracy. Figure 9 shows the trend of accuracy against the number of layers. The accuracy of 70.2% was achieved.

We further optimize the model by varying the number of epochs in

{25, 50, 100, 200, 300, 400, 500}. We find that the accuracy increases with the number of epochs, as shown in Figure 10. However, we observe that the accuracy change is minor after the epochs increase from 300 to 500. The accuracy change is minor compared to the cost of the number of epochs. Thus, we selected the 300 epochs with an accuracy of 78.1 for comparison with other models.

Table 1 shows the Precision, Recall, F1- score, and Support for all the seven emotion types and the average value. The labels 0 represent angry, 1 express disgust, 2 represent fear, 3 represent happy, 4 represent sad, 5 represent surprise, and 6 represent neutral. We observe that the proposed model performs better for each emotion. We trained and reported the results of the other models for the same number of epochs and layers, i.e., five layers and 300 epochs. We tested state-of-the-art models like AlextNet, VGG, GoogleNet, and ResNet, as shown in Table 2. In terms of accuracy, the proposed CNN models outperform existing standard networks. All networks' accuracy is slightly lower because the dataset is small and challenging. We anticipate that the proposed system will have improved accuracy under large datasets training. However, our aim was to design a system with good accuracy under small datasets.

**Table 2:** Comparison of the proposed model with standard networks

| Model | Accuracy (%) |
|---|---|
| AlexNet | 69.6 |
| GoogleNet | 74.1 |
| ResNet | 71.2 |
| VGG | 68.7 |
| CNN | 78.1 |

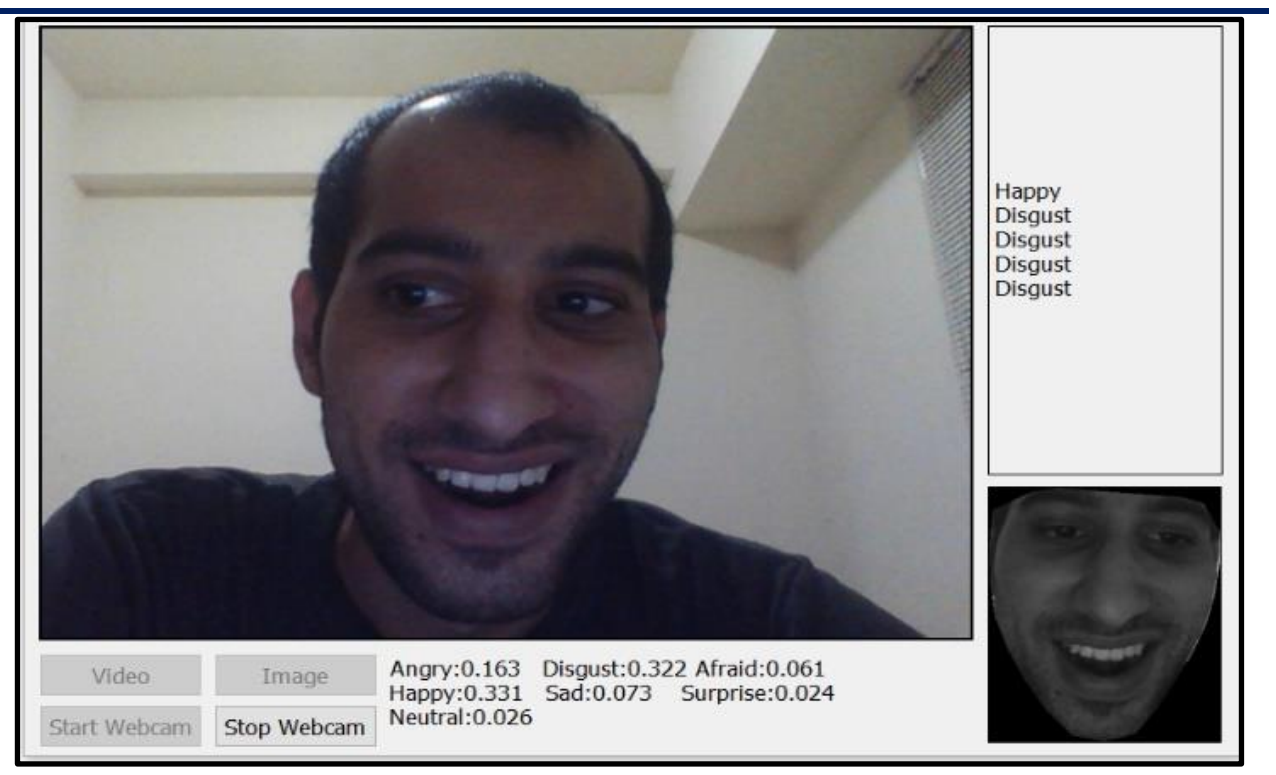

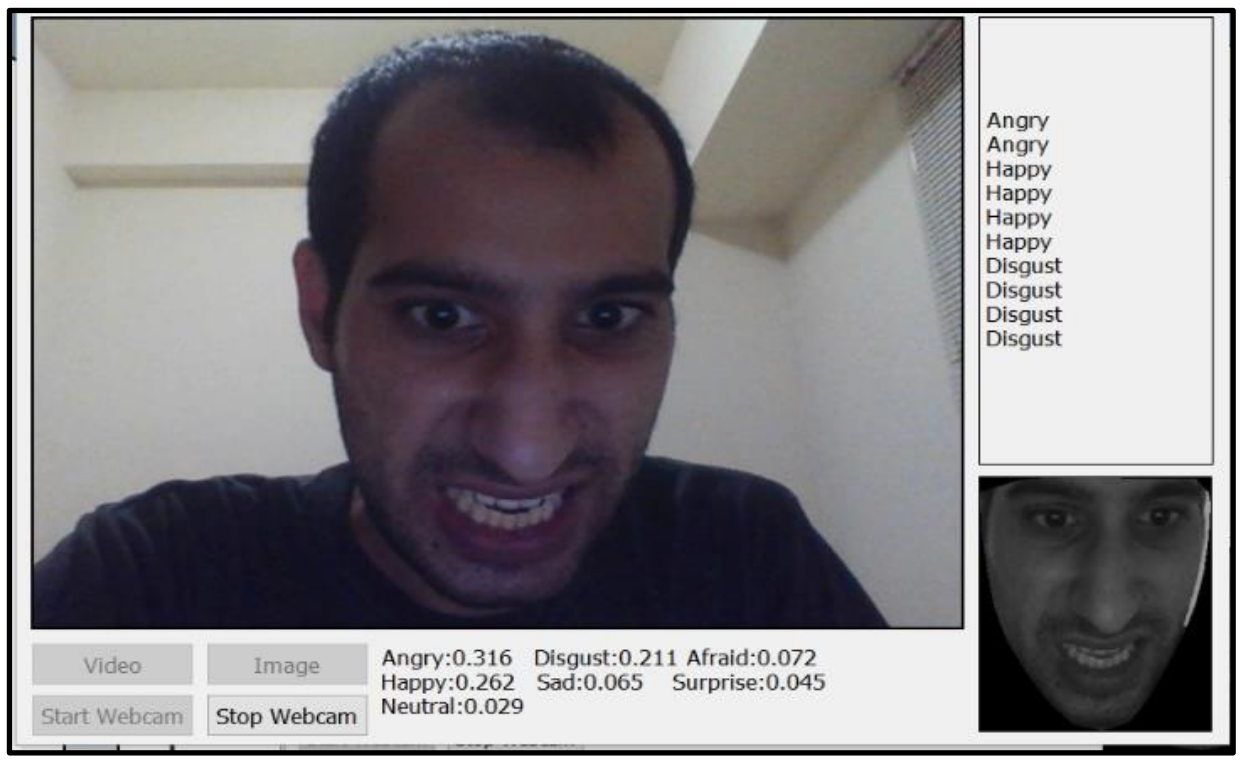

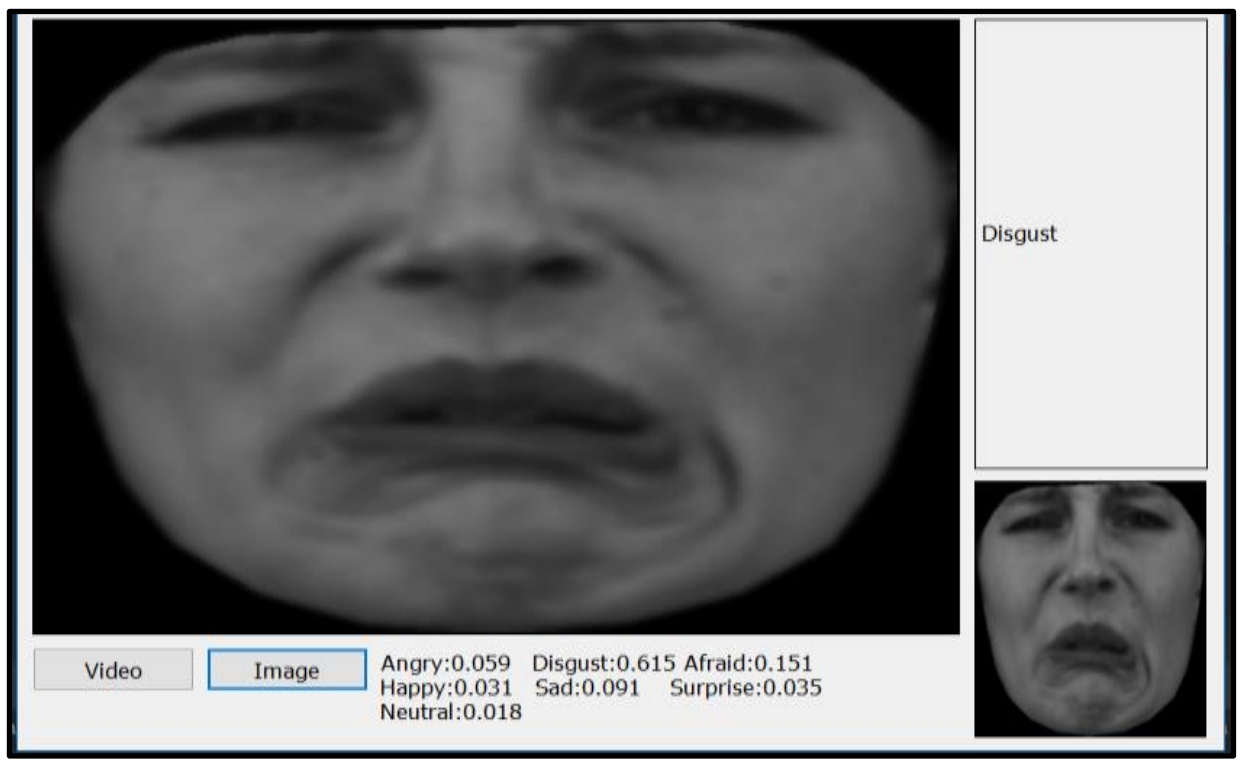

**Figure 11**. Images display in the graphical user interface. (**Top**): a happy face, (**Middle**) an angry face, and (**Bottom**) a disgusted face.

## IV. GRAPHICAL USER INTERFACE

The trained model was then turned into an application. The application developed works in dual mode, which works on still images and live video feeds. The user has to select between each mood. Preprocessing and labeling each image takes about 20-30ms. The interface shows the video's original image or frame and the cropped face image. Two labels have been included in the interface. One label shows the probabilities of all the labels, while another shows the predicted emotion. The average probability of the related feelings of all the images is calculated to increase the chances of proper label selection, i.e., accuracy. Figure 11 (Top) shows a happy face, so the corresponding emotion is shown on the label at the top-right. Figure 11 (Middle) is a continuation frame in which the emotion is angry, showing the anger on the label. Figure 11 (Bottom) shows a disgusting emotion, and the corresponding label is shown on the left.

The probabilities in all three images are displayed at the bottom.

## V. CONCLUSION

In this article, we proposed an efficient system for facial emotion recognition using a convolution neural network, which predicts and assigns probabilities to each facial emotion. Moreover, our proposed system processes each image with diverse preprocessing steps for better prediction, as deep learning models learn from data. The results show that the proposed system performs better compared to other models. We then designed an application of the proposed system with a graphical user interface that classifies emotions in real-time. The results and the live display show that the system has quite a promising performance. Complex network leads to heavy models, which perform poorly during live video. We also observed that although the preprocessing stage is essential, unnecessary preprocessing causes the loss of many features and may lead to poor results. In the future, we aim to further improve the system prediction performance and deploy the proposed system to emotion scanning in a live interview system.